# Computational Induction of Prosodic Structure

## Dafydd Gibbon

## Bielefeld University, Germany

## Abstract

The present study has two goals relating to the grammar of prosody, understood as the rhythms and melodies of speech: to provide an overview and a proposal. First, an overview is provided of the computable grammatical and phonetic approaches to prosody analysis which use hypothetico-deductive methods and are based learned hermeneutic intuitions about language. Second, a proposal is presented for an inductive approach to fill a major methodological gap in the hypothetico-deductive methods: an inductive grounding in the physical signal, in which prosodic structure is inferred using a language-independent method from the low-frequency spectrum of the speech signal. The grammar of prosody is understood here as the autonomous structure of prosodic patterns, the rhythms and melodies of speech, rather than as the relations between prosody and the grammar of words and their combinations.

The overview includes a discussion of computational aspects of standard generative and post-generative models, and suggestions for reformulating these to form inductive approaches. Also included is a discussion of linguistic phonetic approaches to analysis of annotations (pairs of speech unit labels with time-stamps) of recorded spoken utterances. The proposal introduces the inductive approach of Rhythm Formant Theory (RFT) and the associated Rhythm Formant Analysis (RFA) method are introduced, with the aim of completing a gap in the linguistic hypothetico-deductive cycle by grounding in a language-independent inductive procedure of speech signal analysis.

RFA uses spectral analysis of the envelope of speech signals to identify linguistically interpretable rhythm formants (R-formants) in the low-frequency spectrum (LFS) as high magnitude frequency clusters in the very low frequency band below 20 Hz, named by analogy with formants in the high-frequency spectrum (HFS) of vowels. The validity of the R-formant analysis method is demonstrated using the clear case of rhythmical counting, showing clear similarities and differences between the two main R-formants of Mandarin (Standard Chinese) and the three main R-formants of British English, roughly relating to the traditional distinction between syllable and stress timing. The different roles of amplitude and frequency modulation of speech in Mandarin and English story-telling data are analysed for the first time, showing a difference in correlation between the LF AM and FM spectra in the two languages which isrelated to their phrasal and lexical grammatical differences.

The overall conclusions are (1) that normative language-to-language phonological or phonetic comparisons of rhythm, for example of Mandarin and English, are too simplistic, in view of diverse language-internal factors due to genre and style differences as well as utterance dynamics, and (2) that language-independent empirical grounding of rhythm in the physical signal is called for.

# 1   Prosody: data, domains, methods

## 1.1   Prosody and grammar: patterns and functions

Prosody, informally defined in the present context as the rhythms and melodies of speech, is in the meantime one of the most extensively researched areas of spoken language in many disciplines, from phonetics through the sociology and psychology of language to speech technology. It is the domain which most conspicuously distinguishes the grammars of spoken and written language.

The term 'prosodic grammar' is ambiguous, referring (1) to the relation between prosody and the locutionary grammar of lexical items and their combinations at phrase, sentence, text and discourse ranks, or (2) to the autonomous metalocutionary grammar of prosodic patterns alone, independently of the locutionary grammar, as in the finite state intonation grammar of Pierrehumbert (1980) and the finite state tone sandhi grammar of Gibbon (1987, 2001) and Jansche (1998).

The second meaning is adopted in the present study, where 'prosodic structure' refers to systematic low-frequency temporal patterns (rhythms) of the amplitude modulations of speech, and their relation to syllable, foot and phrase domains. The low-frequency temporal patterns (rhythms) of speech melodies, the frequency modulation of speech, are addressed but not the actual forms (melodies) of tone and intonation. The phonetics of tone and intonation have been very thoroughly investigated, but speech timing in the domains of tone and intonation has been less intensively researched than pitch contours in these domains.

The patterns of autonomous prosodic grammar, both in melody and in rhythm, differ not only from language to language (Hirst and di Cristo 1998), and from dialect to dialect, but also in the speech styles and registers of each language and dialect, in everyday conversation and in formal styles of reading aloud and public oratory (Couper-Kuhlen 1993), and also rhythm variation within any given utterance (Gibbon and Li 2019).

Although the patterns of prosody tend to differ in different languages and language varieties, the functions of prosody are relatively similar from language to language. In the grammar and semantics of locutions, lexical tones, as in Mandarin Chinese, lexical pitch accents, as in Japanese, and and lexical stress positions with pitch accent correlates, as in English, all function to distinguish morphemes. In the case of morphological tone and the final tones of intonation patterns, prosodic patterns also have meaningful morphemic status. In the grammar and semantics of prosody as metalocution, pitch accents are deictic morphemes, in that they 'point' to the positions of key words in utterances as metadeictic gestural indices (Gibbon 1983): morphemes and words in temporal utterance locations are the semantic domain of these metalocutions. In writing, highlighting with italics, bold face or underlining have a similar, but spatial metalocutionary deictic function.

Phrasal rhythms and melodies are constructions which consist of metalocutionary morphemes, which 'point' to and disambiguate larger domains of phrase and sentence structure: *She didn't come because she was busy* (she did come – but not for that reason), with pitch and timing patterns contrasting with *She didn't come, because she was busy* (she did not come – and for that reason). At the text and discourse ranks, prosody 'points', among other things, to utterance completeness in incompleteness and parenthesis in narrative and argument patterns, and to patterns of turn-taking in dialogue grammar, such as question (incomplete) and answer (complete) in adjacency pairs.

Methods used in the study of the prosodic structure of these metalocutions have been partly phonological and symbol-phonetic, arguing deductively from linguistic structure to phonetic representations in terms of symbols such as those of the International Phonetic Alphabet. Some methods have been partly signal-phonetic, searching deductively for quantitative physical and physiological correlates of symbols in the speech signal. Typical research questions pertain to isochrony (regular timing) in syllable or foot durations; the alternation or oscillation property of rhythm has generally been ignored. Inductive studies, which use language-independent methods to induce patterns from the speech signal, and only *a posteriori* associate the resulting patterns with linguistically relevant units, are much rarer. Each of these approaches is discussed in this study.

The present contribution focuses on an inductive signal-phonetic methodology for discovering prosodic structure, using methods which are more typical of speech technology than of linguistic phonetics. A detailed overview of formal deductive and inductive methodology in computational phonology and symbol-phonetics is included in order to provide linguistic context. With this background, the inductively oriented *Rhythm Formant Theory* (*RFT*) and its associated methodology of Rhythm Formant Analysis (RFA) are introduced as a very different approach from previous deductive and isochrony-oriented signal-phonetic studies, and a new concept of rhythm formant (R-formant) is introduced.

## 1.2 Open questions: deduction and induction, writing and speeches

There are still many open questions in the study of prosody. The present study calls four basic epistemologically significant principles and widely accepted rationalist practices into question, and suggests alternative empiricist principles and practices:

1. Data types:

    Option 1. The writing bias: a common practice in linguistics is to write down systematically designed sentences and think intuitively about how they would be pronounced. The term 'grammar' has an honourable history as the study of writing, of course (cf. Greek 'γράμμα', 'grámma', letter) and its modern interpretation in the theory of language structure owes much to this history, with the bias which schooling and its cultural normalisation of written text as the canonical form of language expression brings with it.

    Option 2. An alternative is to start with a corpus of actual speech recorded in the wild, or in well-defined, non-elicited scenarios, ranging from specific circumscribed events to random collation: corpus linguistics and corpus phonetics.

2. Domain:

    Option 1. A common, though not universal, linguistic practice in analysing the relation between prosody and grammar is to restrict attention to words and their constituents, and to sentences and their constituents.

    Option 2. An alternative is to include the structures and functions of spoken discourse, including the grammar of interactive dialogue and the text grammar of descriptive, narrative and argumentative monologues within the scope of prosody and grammar.

3. Argumentation:

Option 1. A common practice in linguistic prosody analysis proceeds with deductive logic: abstract premises, in practice written sentences whose component morphemes are attached to a tree structure or an autosegmental lattice as terminal nodes, are mapped to symbol-phonetic theorems by means of phonological rules of inference, for further testing.

Option 2. An alternative is to start with the induction of signal-phonetic representations from speech signals using a combination of quantitative and categorial methods, by analogy with the procedures of automatic speech recognition (ASR). These are then be related to more abstract, linguistically relevant representations. The starting point would be well-established 'universal' principles of acoustic physics and the associated mathematics.

4. Implementation:

Option 1. Commonly, small sets of specific rules and their interaction are examined and phonetic theorems are deduced for further testing against native speakers' intuitions. In some cases, well-defined algorithms are used to interpret the phonological rules consistently.

Option 2. An alternative is compute patterns from recorded data in an inductive procedure, and create comprehensive models which are both theoretically and empirically sound and complete with respect to both formalism and data. This is a procedure which is more typical of speech technology and natural language processing, and uses pattern matching, classification and prediction algorithms.

The present study follows Option 2 in each of these methodological categories of Data, Domain, Argumentation and Implementation, within the constraints of RFT and the relevant data.

## 1.3 Overview

After the methodological overview in Section 1, Section 2 provides a brief overview of deductive and inductive phonological and symbol-phonetic theories. Section 3 is concerned with specific deductive signal-phonetic approaches and Section 4 introduces the inductive approach to signal-phonetic prosodic analysis and its extension in Rhythm Formant Theory (RFT) and Rhythm Formant Analysis (RFA). Section 5 applies RFA to the analysis of Rhythm formants (R-formants) in Mandarin story reading, and Section 6 summarises, concludes and suggests further developments. After the references an appendix is provided with the code which was used to calculate selected computational examples in the text, in the interest of encouraging further study.

## 2 Deduction and induction in computational phonology

### 2.1 Stress reduction, metrical tree, autosegmental tiers and finite state transduction

In view of the internal variability of languages in terms of dialects, styles, genres, rhetorical strategies, and the dynamic variability of individual utterances, comparisons of entire languages as homogeneous constructs are far from being empirically grounded, as relevant situational variables are not controlled for. In studies of grammar, this variability issue is generally side-stepped by relying on normative hermeneutic judgments about hypotheses which are derived deductively, top-down, from a theory and grounded in the intuitions of the linguist as an 'idealised native speaker'. Similarly, in phonological and phonetic studies of prosody, both of rhythm and of melody, language-internal variability has also been largely ignored, for example in establishing singular timing regularity indices of around 40 and 65 for Mandarin or for English, respectively. Such results are empirically meaningless without careful control of the variability factors.

Normative stipulation of this kind is common practice but it is not an empirical procedure. In an empirical procedure, the incomplete hypothetico-deductive cycle of rhythm modelling needs to be grounded in articulated, transmitted and perceived physical signals from which structures to match top-down hypotheses are inferred inductively, bottom-up.

Inducing linguistically relevant structure from the speech signal is the characteristic task of automatic speech recognition. However, the Hidden Markov Model and Neural Network algorithms which are typically used are opaque black boxes from a linguistic point of view, and the linguist cannot find linguistically interesting units or structures in them. As a step towards ensuring transparency, before proceeding to a inductive quantitative signal-phonetic methodology two classic symbol-phonetic deductive algorithms are discussed as background context: the stress subordination algorithm of Chomsky and Halle (1968) and the metrical algorithm of Liberman (1975) and Liberman and Prince (1977). These algorithms map syntactic tree structures (whether as tree graphs or as bracketings is immaterial) to sequences of numbers which are interpreted as patterns of degrees of stress. Arguably the most well-known examples of the deductive approach to prosody and grammar are:

1. stress subordination theory (Chomsky et al. 1956; Chomsky and Halle 1968) and the metrical theory (Liberman 1975; Liberman and Prince 1977) theory of stress patterning;

2. autosegmental (Goldsmith 1990) and finite state (Pierrehumbert 1980; Gibbon 1987, 2001; Kay 1987) theories of independent parallel prosodic patterns.

3. optimality theory, which re-interprets the deterministic inference approaches of the stress subordination, metrical, autosegmental and finite state theories as a heuristic search problem with potentially more than one solution within the search space of phonological patterns, therefore in principle non-deterministic.

In the present context, the main question is whether these deductive systems are suitable for approaches to speech data analysis in which models are derived inductively and related *post hoc* to linguistic categories of syllable, foot, word, phrase, sentence, or higher ranks such as text and dialogue: the *Rule Reversibility Question* (*RRQ*).

The *RRQ* has been posed from time to time in the context of generative phonology during investigations of the relation of phonological rules to the production and perception of language in performance. The simple answer to the *RRQ* is positive, and is shown below for linear phonological rules as well as for hierarchical stress subordination, metrical, autosegmental and finite state approaches.

For optimality-theoretic approaches, which are orientated towards finding specific cases by narrowing a search space rather than generating specific cases, the *RRQ* is not quite so clear. Since it has been shown that the essential features of optimality theory can be formalised with finite state transducers (Karttunen 1998), and these are reversible (see the following subsections), it is not necessary to discuss optimality theory in the present context.

## 2.2  RRQ: reversibility of phonological rules

Basic phonological rules are of the type γ α δ → γ β δ, abbreviated as α → β / γ _ δ, where α, β γ and δ are either feature structures or phonemes, morphophonemes, archiphonemes or other phoneme-like segments. Basic phonological rules have been shown in Finite State Phonology to be equivalently formalisable as finite state transducers (finite state automata with an input and an output), which can in turn be composed into a comprehensive finite state system (Johnson 1972; Koskenniemi 1984; Kaplan and Kay 1994; Carson-Berndsen 1997). Phonological rules are in general deductive and top-down, unidirectional from phonology to phonetics, and deterministic.

But is also known that finite state transducers are reversible. This means, fortunately, that phonological rules are also reversible, that is, α → β / γ _ δ if and only if β → α / γ _ δ (if rule ordering is also reversed). So in principle an inductive phonology proceeding from phonetic data to phonological generalisations is formally possible. Autosegmental rules can be formalised as multi-tape finite state transducers (Kay 1987); consequently an inductive autosegmental phonology is also possible, as shown by Berndsen (1998).

However, a transducer which is deterministic in the top-down direction may be non-deterministic, i.e. ambiguous, in the bottom-up phonetics-to-phonology direction, and therefore requires additional context for disambiguation. For example, the Mandarin tonal sandhi rule

*Tone3 → Tone2 / ___ Tone3*

transduces a lexical sequence *Tone3^Tone3* into the phonetic sequence *Tone2^Tone3*. Reversing the rule yields two lexical results, *Tone2^Tone3* or *Tone3^Tone3*, because a lexical sequence *Tone2^Tone3* may also occur:

Tone2^Tone3 → {Tone3, Tone 2} / _ Tone3

For example (disregarding the controversy about the extent to which Tone 3 sandhi is due to phonological or phonetic factors, the minor differences in rhythm, and the difference in morphological structure) the phonetic representation *mei2jiu3* in a reverse analysis is ambiguous:

美酒 *mei3jiu3*: (tasty wine) ~ 梅酒 *mei2jiu3*: plum wine

Consequently, a reversal of the rule results in ambiguity and disambiguation depends on the syntactic, semantic or pragmatic context

## 2.3  RRQ: reversibility of Niger-Congo tone sandhi rules

Gibbon (1987, 2001) showed that tone sandhi can be represented as a finite state transducer with two levels (traditionally: 'tapes') which unites the separate contexts of standard phonological rules such as H → !h / L __ ('a high tone is realised as downstepped high after a low tone') into a coherent connected system. Gibbon (2018) showed that the phonetic interpretation can be extended by a third stage with numerical functions (making the second stage redundant), which can then in turn be operationally tested in a speech synthesiser (Figure 1).

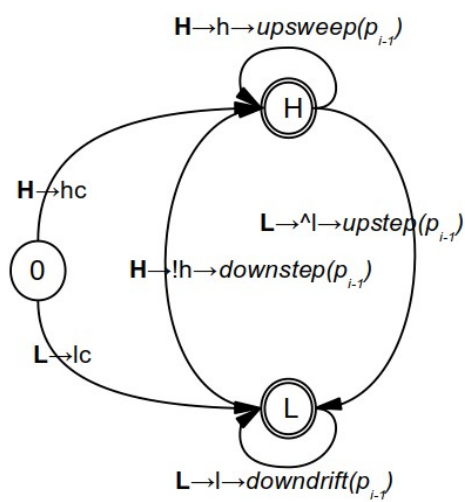

*Figure 1: Three-stage interpretation of phonetic tone sandhi in a subsystem of a two-tone Niger-Congo language with a three-tape finite state transducer.*

Figure 1 represents the core finite state grammar of tonal sandhi in a Niger-Congo two-tone language type such as Tem (ISO 639-3 *kdh*; Tchagbalé 1984), abstracting away from some contextual details. Node labels represent the dual transduction function: $\mathbf{H} \rightarrow !h \rightarrow downstep(p_{i-1})$, for example, represents a transduction of a lexical high tone, $\mathbf{H}$, into a downstepped symbol-phonetic tone, !h, followed by a second transduction into a numerical signal-phonetic F0 (fundamental frequency, F0, 'pitch') functions which derive a pitch value $downstep(p_i)$ from the previous signal-phonetic F0 pitch value $downstep(p_{i-1})$. The intermediate stage of the transducer shown in Figure 1 can omitted: the abstract lexical tone can be mapped directly to a physically interpretable numerical value. Each transition between two nodes in the transducer is, in a full model, associated with a variable for syllables bearing the tone concerned, thus implementing the autosegmental relation of tone-text association.

Tem, and many other Niger-Congo tone languages, have considerably more complex tone sandhi, but this grammar captures the basic principle. The first stage of the phonetic interpretation is represented by the mapping of a high or low lexical (phonological) tone $\tau_{\text{lexical}} \in \{\mathbf{H}, \mathbf{L}\}$ to a phonetic tone $\tau_{\text{phonetic}} \in \{h, l, !h, ^l\}$, I.e. high, low, downstepped high, upstepped low. Language-specific constraints such as tone-blocking consonants can also be specified. As with other phonological finite state transducers, the transducer is reversible, thus opening the possibility of using it as a model for components of automatic speech recognisers.

Jansche (1998) has adapted this approach to the modelling of the Tianjin tone sandhi. Applications of this computational approach to standard Pǔtōnghuà (Mandarin), which would be rather straightforward, and to other Chinese dialects or Sino-Tibetan languages are not available.

## 2.4   RRQ: reversibility of the stress cycle

First, the well-known stress cycle algorithms of Chomsky and Halle (1968) and of Liberman (1975) and Liberman and Prince (1977) are outlined. These algorithms deduce numerical terminal node sequences from tree structures, and the numerical sequences are in turn interpreted as the prominence patterns associated with different stress positions. Then a previously unpublished inverse algorithm for mapping number vectors to tree structures, rather than tree structures to numbers, is formulated.

The example of a stress cycle discussed here is the core of the *Nuclear Stress Rule* of Chomsky and Halle (1968), which stipulates that the right hand constituent of a binary tree or subtree is more strongly stressed (indexed with a lower number) than the left-hand constituent. The same principle applies to the core of the *Compound Stress Rule*, where it is the left-hand constituent which is more strongly stressed. Then, Then a symbol-phonetic inductive algorithm which performs the inverse operation of mapping a sequence of numbers on to a tree structure is introduced as the inverse of the stress subordination and metrical procedures. Inverse algorithms of this type have not previously been published in this context (but cf. Gibbon 2003 for an outline in a different context).

The stress cycle is problematic in a number of ways. First, when the computed stress value exceeds about 4, native speaker judgments fail; to counter this, readjustment rules were proposed. Second, in discourse contexts, other factors such as contrast, emphasis, and rhetorical effect co-determine stress levels.

In the literature, the terminology used to describe stress and its correlates is unfortunately very inconsistent. To clarify, the usage in the  present study is as follows:

1. Crucially, *stress is a position in a linear or hierarchical structure*. In this role, stress has no intrinsic phonetic content, unlike lexical tone or lexical pitch accent in other language types, but may be expressed with a variety of pitch patterns. The stress subordination, metrical and parenthesis counting algorithms make the positional character of stress in English very clear.

2. Stress positions in words are determined lexically, but can be modified in phrasal contexts, and these can in turn be modified in rhetorical discourse contexts.

3. Perceptually, abstract positional stress is heard as *prominence* of constituents at a given stress position. The prominence percept is quite simply a metalocutionary pointer to a temporal location in the locutionary word sequence.

4. The prominence percept is a function of perceived pitch height, pitch change and timing, caused in speech production by complex interaction of *articulatory effort*, *phonation rate* and *clarity of speech sounds*. In acoustic transmission the prominence percept is due to a complex function of *fundamental frequency* (F0, often misnamed 'pitch'), *duration* and *amplitude* constraints.

The core of the bottom-up stress subordination algorithm of Chomsky and Halle (1968) for the nuclear stress rule applies to tree structures in the form of parenthesis notations:

1. Assign all terminal (lexical) items the stress index 1.

2. Repeat until all brackets have been removed:

    2.1 Reduce stress values in the innermost brackets by 1, except the right-hand value.

    2.2 Remove the innermost brackets.

Table 1 shows the stress value assignments for the simple case *big John saw Tom's dog*.

*Table 1: Deductive cyclical stress assignment by stress subordination (Chomsky and Halle 1968).*

| Cycle inputs and outputs | Rule application | Cycle |
|---|---|---|
| ((big John) (saw (Tom's dog))) | Input | |
| (($^1$big $^1$John) ($^1$saw ($^1$Tom's $^1$dog))) | by Rule 1 | |
| (($^2$big $^1$John) ($^1$saw ($^2$Tom's $^1$dog))) | by Rule 2.1 | 1$^{st}$ cycle |
| ($^2$big $^1$John ($^1$saw $^2$Tom's $^1$dog)) | by Rule 2.2 | |
| ($^2$big $^1$John ($^2$saw $^3$Tom's $^1$dog)) | by Rule 2.1 | 2$^{nd}$ cycle |
| ($^2$big $^1$John $^2$saw $^3$Tom's $^1$dog) | by Rule 2.2 | |
| ($^3$big $^2$John $^3$saw $^4$Tom's $^1$dog) | by Rule 2.1 | 3$^{rd}$ cycle |
| $^3$big $^2$John $^3$saw $^4$Tom's $^1$dog | by Rule 2.2 | |

Rather than using discrete stress values, Liberman (1975) and Liberman and Prince (1977) defines a relational concept of stress, with the same kind of input as the stress subordination algorithm. The top-down metrically based algorithm can easily be checked by inspecting Figure 2.

1. Assign *w* to each left branch and *s* to each right branch.

2. For each terminal node:

   1. Move up the branch and, beginning at the first *w*, count each higher node on the branch, including the *r* node.

   2. Assign the resulting number to the terminal node as stress index.

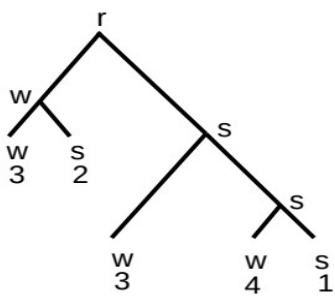

*Figure 2: Metrical tree (Liberman 1975).*

There is a simple deductive top-down left-right deterministic linear parenthesis counting algorithm which achieves the same result as the stress subordination procedure. This algorithm treats the input bracketing as a string in a parenthesis language, in terms of formal grammar (Knuth 1967), in which the brackets do not simply mark structure but are part of the vocabulary. The algorithm initialises a counter and then iterates through the string (cf. Code Appendix, Section 9.2):

1. Remove the outer brackets and initialise the counter to 1.

2. From left to right:

   1. if a left bracket is encountered, add 1; if a terminal element is encountered after the left bracket, annotate the terminal element with the current counter value

2.  if a right bracket is encountered, subtract 1; if a left bracket or the end of the string is encountered after a right bracket, annotate the previous terminal element with the current counter value.

The algorithm can be implemented as a finite state transducer enhanced with a parenthesis counter indicating stack positions without actually using a stack. The automatically generated output of a well-formed bracketed sequence "( ( *tiny Moll* ) ( *met* ( *tall Jill* ) ) )" is shown in Figure 3; cf. also Section 9.2, Code Appendix).

| | |
|---|---|
| `Counter: 1` | `Input: ( Output: []` |
| `Counter: 2` | `Input: ( Output: []` |
| `Counter: 3` | `Input: tiny Output: []` |
| `Counter: 3` | `Input: Moll Output: ['3tiny']` |
| `Counter: 3` | `Input: ) Output: ['3tiny', 'Moll']` |
| `Counter: 2` | `Input: ( Output: ['3tiny', 'Moll']` |
| `Counter: 3` | `Input: met Output: ['3tiny', '2Moll']` |
| `Counter: 3` | `Input: ( Output: ['3tiny', '2Moll', '3met']` |
| `Counter: 4` | `Input: tall Output: ['3tiny', '2Moll', '3met']` |
| `Counter: 4` | `Input: Jill Output: ['3tiny', '2Moll', '3met', '4tall']` |
| `Counter: 4` | `Input: ) Output: ['3tiny', '2Moll', '3met', '4tall', 'Jill']` |
| `Counter: 3` | `Input: ) Output: ['3tiny', '2Moll', '3met', '4tall', 'Jill']` |
| `Counter: 2` | `Input: ) Output: ['3tiny', '2Moll', '3met', '4tall', 'Jill']` |
| `Counter: 1` | `Input: [] Output: ['3tiny', '2Moll', '3met', '4tall', '1Jill']` |

*Figure 3: Derivation of stress number production from a bracketed string by means of a finite state transducer enhanced with a stack-like parenthesis counter. Implementation: Python, cf. Section 9.2, Code Appendix.*

Each of these three algorithms, stress subordination, metrical and parenthesis counting, defines just one subset of a range of a possible stress patterns which can be intuitively confirmed by native speakers.

Gibbon (2018) noted that stress cycle rules are reversible, permitting the induction of a metrically labelled syntax tree from a sequence of numbers. The associated words are omitted in the interest of brevity in the illustration of a bottom-up inductive procedure, which uses a standard shift-reduce parser to implement the inverse of the stress reduction and metrical algorithms (cf. Table 2).

*Table 2: Automatic bottom-up (shift-reduce) parse of input number vector (top row) to a number-labelled tree structure (bottom row). Implementation: Python.*

| INPUT | STACK |
|-------|-------|
| **Start:** [[3], [2], [3], [4], [1]] | →0 : [] |
| **Shift:** [[2], [3], [4], [1]] | →1 : [[3]] |
| **Shift:** [[3], [4], [1]] | →2 : [[2], [3]] |
| **Reduce:** | ⇐1 : [[[3, 2], 2]] |
| **Shift:** [[4], [1]] | →2 : [[3], [[3, 2], 2]] |
| **Shift:** [[1]] | →3 : [[4], [3], [[3, 2], 2]] |
| **Shift:** [] | →4 : [[1], [4], [3], [[3, 2], 2]] |
| **Reduce:** | ⇐3 : [[[4, 1], 1], [3], [[3, 2], 2]] |
| **Reduce:** | ⇐2 : [[[3, [4, 1], 1], 1], [[3, 2], 2]] |
| **Reduce:** | ⇐1 : [[[[3, 2], 2, [3, [4, 1], 1], 1], 1], 1]] |

The shift-reduce parsing algorithm (Aho et al. 2006) uses a stack to store intermediate results, shifting elements from the input on to the stack and at each stage attempting to reduce the stack by combining adjacent elements into a tree structure until, for well-formed sequences, the stack only contains a single tree. The reduce criterion is that the number at the top of the stack is greater than the number of the next item on the stack. In order to make this possible, the lower input number is attached to the tree created by the reduce operation. For example, an input sequence [[3], [2]] appears on the stack in reverse order as [[2], [3]] ([3] is at the top of the stack), and is reduced to a tree structure [[[3, 2], 2]] with the lowest number, 2, attached to the tree structure.

There is also a simple equivalent inverse algorithm for parsing a sequence of numbers into a string in a parenthesis language. Starting with an initial seed value of 1, if the previous number is smaller, subtract it from the current number and output that number of left brackets, otherwise subtract the current number from the previous number and output that number of right brackets. A derivation is shown in Table 3 (cf. Section 9.3, Code Appendix).

*Table 3: Parsing of a string in a parenthesis language from a sequence of numbers - inverse 'Nuclear Stress Rule'. Implementation in Python. See Section 9.3 Code Appendix.*

| | |
|---|---|
| Input: | [3, 4, 2, 3, 4, 1] |
| Intermediate outputs: | (( 3 |
| | (( 3 ( 4 |
| | (( 3 ( 4  2 )) |
| | (( 3 ( 4  2 ))( 3 |
| | (( 3 ( 4  2 ))( 3 ( 4 |
| Final output: | (( 3 ( 4  2 ))( 3 ( 4  1 ))) |

# 3    Signal-phonetic deduction

## 3.1    Annotation-based isochrony metrics

Discussions of speech rhythm in phonetics and linguistics during the past fifty years have been mainly deductive in the sense outlined in Section 1, like the phonological approaches outlined in Section 2. The deductive approaches to rhythm description in phonetics have been diverse and controversial, and have not always addressed core features of standard characterisations of rhythm as alternation or oscillation. The present subsection addresses the formal foundations of such analyses and the following subsection discusses selected examples of their application.

In phonetics, the deductive approaches started with simple categorial distinctions between mora, syllable and foot timing, then progressed to quantitative scales. The most popular quantitative approaches for over half a century have measured relative *isochrony* of speech segments, that is, the degree to which sequences of phonological units such as consonantal and vocalic speech intervals, syllables or feet have similar durations (Roach 1982; Jassem et al. 1984; Scott et al. 1985 and many later studies using variance of consonantal durations and percentages of vocalic durations, cf. Dellwo and Wagner 2003). The simplest of these isochrony measures is variance (or standard deviation), and, like variance, the others are also variants of measures of dispersion around the mean duration. It is evident that such global dispersion measures are not models of oscillating rhythms but heuristic indices of relative evenness of duration: evenness of timing is a consequence of regular alternation or oscillation, but may also occur without alternation: the same standard deviation value, for example, is achieved by all possible orderings of a given sequence.

The most widely used and the most successful of these measures is the *Pairwise Variability Index*, (*PVI*), introduced by Low et al. (2000) as local dispersion measure which reduces the influence of variation in speech rate. The *PVI* may be better understood by interpreting it as a distance measure between adjacent items. A non-normalised 'raw' version, *rPVI*, is typically used for consonantal utterance chunk sequences, whose duration is relatively invariant, and normalised version, the *nPVI*, is typically used for vocalic chunk sequences, syllables and feet, which tend to vary as a function of changes in speech rate. The two *PVI* variants average the differences between adjacent values in a vector of durations $D = (d_1, \dots d_n)$ and are standardly formulated over vectors of consonantal, vocalic, syllabic, etc., durations in annotations of the speech signal:

$$rPVI(D) = \sum_{k=1}^{n-1} |d_k - d_{k+1}| / (n-1)$$

$$nPVI(D) = 100 \times \sum_{k=1}^{n-1} \frac{|d_k - d_{k+1}|}{(d_k + d_{k+1})/2} / (n-1)$$

The *rPVI* and *nPVI* can be analysed as versions of the basic *Manhattan Distance* metric and its normalised form, the *Canberra Distance* metric, respectively:

$$Manhattan\ Distance : MD(P, Q) = \sum_{k=1}^{n} |p_k - q_k|$$

$$Canberra\ Distance : CD(P, Q) = \sum_{k=1}^{n} \frac{|p_k - q_k|}{p_k + q_k}$$

In the reformulation of the metrics as distance measures, the distance between the subvectors $P = (d_1, \dots d_{n-1})$ and $Q = (d_2, \dots d_n)$, i.e. time-shifted subvectors of the vector $D$, is calculated. In the *rPVI* the sum is averaged, unlike the Manhattan Distance, and in the *nPVI*, in contrast to the Canberra distance, the sum is not only averaged but multiplied by 100. The *nPVI* denominator is also averaged, yielding a nonlinear scale with asymptote of 200 (not a percentage scale as claimed in the literature).

### 3.2 The limits of isochrony measures

The *PVI* measures measure have several drawbacks as measures of rhythm (Gibbon 2003). Nolan and Jeon (2014) addressed some of these points but did not refute them.

A formal issue is that, in contrast to the open-ended linear scale of the *rPVI*, the *nPVI* is a non-linear asymptotic scale with an asymptote of 200 (not a percentage scale, as claimed in the literature), of which only a quasi-linear section below about 100 is empirically useful. The linear and asymptotic variants are clearly numerically incommensurable, though they define the same ranking and the *nPVI* is quite close to linear in the relevant sub-scale (Figure 4). Another formal issue is that the *PVI* variants assign the same indices to variants with alternations (i.e. rhythmical sequences) as to utterances in which adjacent duration differences are arbitrarily positive or negative (i.e. non-rhythmical).

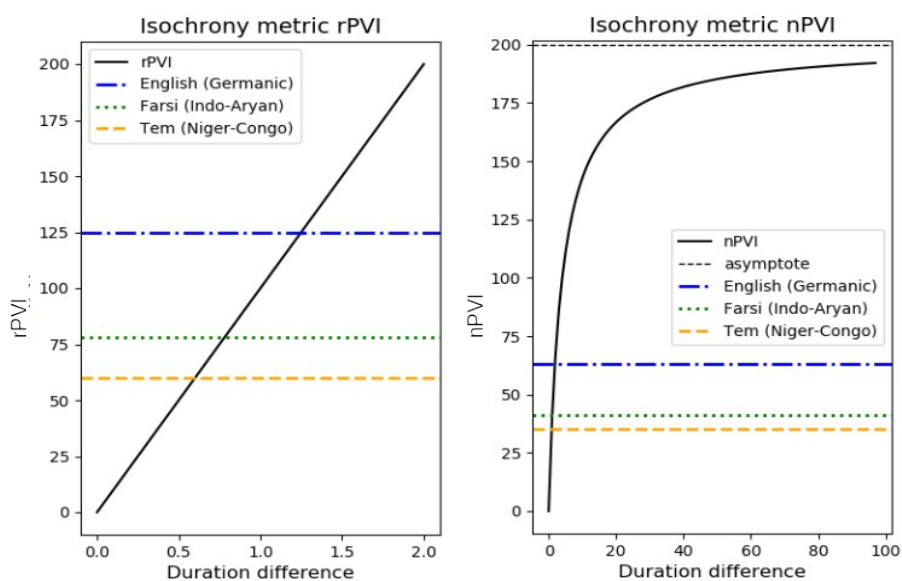

*Figure 4: Illustration of rPVI and nPVI functions with three language samples. Note that the nPVI scale is not a linear percentage scale, contrary to claims in the literature.*

A trivial error of interpretation found in the literature is the claim that ($n$-1) is used to reduce the effect of final lengthening in utterances, where in fact it simply accounts for the number of differences between adjacent items in a sequence of length $n$, which is $n$-1.

A more important empirical issue is that, although they are often termed 'rhythm metrics', in fact in each *PVI* variant removes all rhythmic alternations by taking the absolute value of the subtraction operation. The *PVI* variants are thus measures of relative equality of durations in $D$, not of rhythm, whether alternating or not: they are not rhythm metrics but *isochrony metrics*. To illustrate: it is easily verified that for an alternating 'rhythmic' sequence such as (2,4,2,4,2,4) and a non-alternating, 'non-rhythmic' geometrical sequence (2,4,8,16,32,64), $nPVI$ = 66.67 in each case, and that for alternating (2,4,2,4,2,4) and non-alternating linear (2,4,6,8,10,12), $rPVI$=200 in each case.

Another empirical problem with the *PVI* variants is that the measures are binary, whereas rhythms may be unary, ternary or more complex, patterns which are beyond the capability of the isochrony metric (cf. also Kohler 2009; Tilsen and Arvaniti 2013). The binarity of the variants is evident not only from use of the binary subtraction relation but also from the property of the *PVI* variants interpreted as distance measures between two subvectors of the vector *D*, with the second vector shifted one position in relation to the first.

Wagner (2007) also demonstrated the binarity of the approach by representing the relation between the shifted vectors (as defined above) in a two-dimensional scatter plot, which shows the distribution of this binary relation, rather than a one-dimensional matric value. The plot is constructed by obtaining the *z*-scores of the data, so that the mean appears as zero and different data sets can be compared, and then plotting every adjacent syllable pair with the first syllable on the *x*-axis and the second on the *y*-axis, so that pairs of longer syllables appear top right, pairs of shorter syllables appear bottom left, shorter-longer pairs appear top left and longer-shorter pairs appear bottom right. The scatter plot in Figure 5, for example, clearly visualises differing syllable duration distributions in Farsi (right) as more syllable timed (evenly distributed around the mean), and in English (left) as less syllable timed (skewed towards the bottom left quadrant).

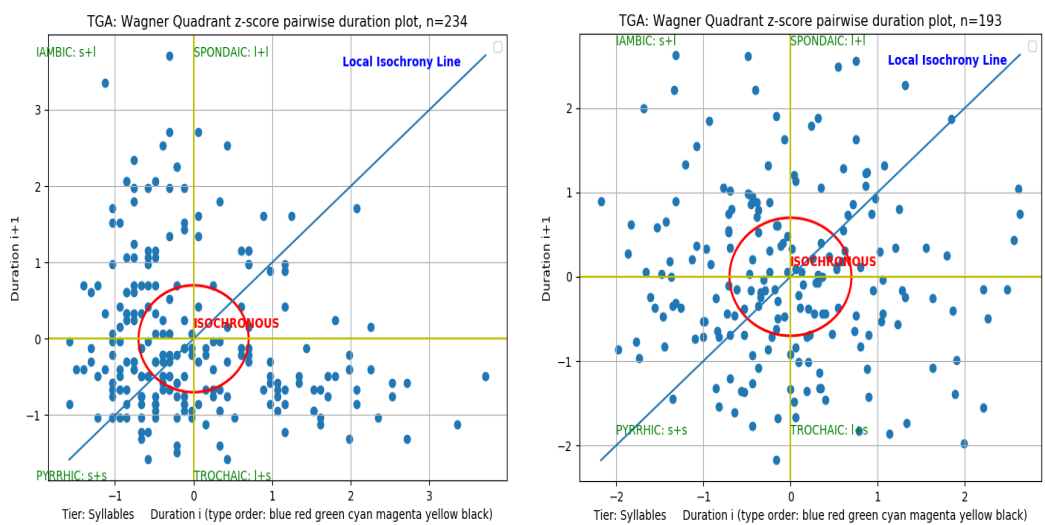

*Figure 5: Wagner scatter plot quadrants, illustrating clear duration differences between shorter-shorter syllable pairs, bottom left quadrant) and other pairs for English (left, news-reading from the Aix-Marsec database, Auran et al. 2004), and more even duration distributions, i.e. more isochronous durations, for Farsi (story reading, from Marzban 2015); reading aloud speech styles.*

The *PVI* measures therefore therefore turn out to be neither empirically 'complete', since they define a binary relation, while rhythms may be more complex, nor empirically 'sound', since they also measure non-rhythms. Like other isochrony measures, the *PVI* variants are formally not *models* of rhythm. Nevertheless the *PVI* measures have been rather succesful as *heuristics* for distinguishing different language types (albeit mainly because the data samples which are typically selected tend to be 'well-behaved' as predominantly alternating and binary, not because the measures inherently distinguish rhythmic from non-rhythmic utterances).

# 4   Signal-phonetic induction

## 4.1   The low-frequency spectrum in production and perception models of rhythm

Parallel to the development of the deductive isochrony models, inductive methods were developed which, in contrast to the annotation-based deductive approaches, take the speech signal directly as input and start by modelling rhythms as oscillating modulations of the amplitude of the speech signal, without reference to linguistic categories. In a posterior step, the results are relatable to annotations of speech sounds, syllables, words and larger units. The oscillating modulation approaches are, formally, *theories* with interpretations as neurobiological *models* (Ding et al. 2017).

Two directions in the inductive, signal-oriented approach developed in parallel: speech production theories and speech perception theories. The production theories postulate a carrier signal, (the fundamental frequency, produced by the larynx) with regular oscillating amplitude modulations (AM) with superimposed consonantal noise and the filter functions of vowel vocal tract shapes (Cummins et al. 1999; O'Dell and Nieminen 1999; Barbosa 2002; Inden et al. 2012). An appropriate procedure for modelling speech rhythm production in the low-frequency spectrum (LFS) is Fourier Synthesis.

Related models of speech perception were independently developed using demodulation of AM oscillations with a variety of procedures to extract the amplitude envelope modulation (AEM) from relatively long segments of the signal, usually >3 s:

1. application of the Hilbert transform, the standard formal method;
2. rectifying and low-pass filtering the signal, the standard practical procedure;
3. peak-picking in a moving window over the rectified (absolute) signal, and low-pass filtering (Gibbon 2018);
4. extraction of the intensity trace from the squared signal  (Dogil and Braun 1988);
5. binary re-scaling of the energy in short-term spectra as a model of sonority in speech (Galves et al. 2002; Fuchs and Wunder 2015).

After demodulation of the AEM, a Fourier transform is applied to determine the LFS of the selected segment of speech, of which a very low frequency segment <20 Hz, often <16 Hz or <10 Hz is used in order to determine spectral peaks which are identified as the frequencies of speech rhythms. In some studies, the signal is first separated into separate lower, mid anFd higher frequency bands and spectral analysis is applied separately to these bands, with results averaged and binned for identifying spectral peaks. In the present study, the signal is low-filtered, ignoring high-frequency bands, and the envelope is extracted by peak-picking in a moving window; after spectral analysis the spectral peaks, or clusters of high magnitude spectral frequencies, are interpreted linguistically and referred to as rhythm formants (R-Formants) rather than simply as spectral peaks.

Many studies using amplitude demodulation approaches have been made of the acoustics of speech rhythms, in neurobiology and musicology (cf. Ding et al. 2017), as well as in phonetics (Todd and Brown 1994; Cummins et al. 1999; Ludusan et al. 2011; Varnet et al. 2017; Tilsen and Johnson 2008; Tilsen and Arvaniti 2013; He and Dellwo 2016; Gibbon 2018; Gibbon and Li 2019;

Suni et al. 2019; Wayland et al. 2020). A common result in many studies is that the major frequency peak in the speech LFS is around 5 Hz. Some studies have found a secondary peak around 2 Hz, which is also found in music (Ding et al. 2017). From a linguistic point of view, the 5 Hz peak (R-formant) relates to the articulation rate of syllables as approximately 200 ms units, and the 2 Hz cluster (R-formant) relates not only to musical bars but also to the articulation rate of approximately 500 ms length foot or word units.

Studies of AM demodulation with spectral analysis have tended to use elicited 'laboratory' data (with the exception of Ding et al. 2017, who used large corpora of speech and music), and have also tended to focus on combining spectral vectors from several frequency bands as indicators of voice quality for clinical phonetic diagnosis or, in phonetic rhythm typology, on lower frequencies as indicators of differences between languages.

The aims of the present study, in contrast, are to apply Rhythm Formant Theory (RFT) to include spectral analysis of frequency modulation (FM) in order to investigate the contribution of the fundamental frequency (F0, 'pitch') to speech rhythms, in particular whether the FM envelope LFS (FEMS) correlates with the AM envelope LFS (AEMS), and whether correlation values depend on variations in language, genre and gender.

## 4.2   Rhythm Formant Theory (RFT)

*Rhythm Formant Theory* (*RFT*) is a development of the inductive LFS analysis approach. RFT makes the following explicit and linguistically informed assertions:

1. Modulation. Speech rhythms are observable as low frequency oscillations in the amplitude and frequency modulations of speech, with measurable frequencies, and are tendentially *a fortiori* isochronous.

2. Rhythm formants. Rhythm formants (R-formants) are higher magnitude clusters of frequencies in the low-frequency spectrum of speech, and are related to linguistic units and to neurobiological events, complementarily to direct application in clinical and speaker or language recognition contexts.

3. Amplitude and frequency domains. R-formants are found both in the amplitude envelope modulation spectrum (AEMS) and in the modulation of the frequency envelope, fundamental frequency, F0, 'pitch' (FEMS).

4. Simultaneous rhythm formants. The R-formants occur in different but overlapping frequency zones, related to the articulation rate of speech units from discourse to phone, and are variable within the zones.

5. Serial rhythm formant changes. Rhythm zones vary with time during discourse, with shifting frequency ranges due to the shifting speech rates of sub-syllabic, syllabic or larger units.

6. Asymmetrical rhythm. In linguistic analyses, rhythms are considered to differ between 'trochaic' and 'iambic' patterns, and RFT employs a dedicated spectrum analysis method to identify physical strong-weak and weak-strong rhythm patterns which are interpretable as linguistic patterns.

The present RFT approach shares point 1 above, and to some extent point 2, with previous LFS-based approaches. Points 3, 4, 5 and 6 are innovations. Points 1, 2 and 3 are dealt with in the present study, and points 4, 5 and 6 are works in progress.

It may be suggested that the AEMS and FEMS generally correlate strongly in formal speech, while correlations may be lower in less formal speech. It may also be suggested that in tone languages such as Mandarin, the correlation is lower, because of the tonally determined, relatively arbitrary changes in F0 from one syllable to another, while in a stress language like English the pitch patterns at adjacent stress positions tend to remain fairly similar until the final stress of the sequence. These differences are investigated in the following sections.

## 5  Induction of rhythm structure in amplitude and frequency modulation

### 5.1  Rhythmical counting: calibration of the method with clear cases

The RFT analysis method is illustrated with the rhythmically clear case of regular fast counting from one to thirty in English (Figure 6), in which a cluster of the most prominent frequencies occurs between 4Hz and 4.5Hz, constituting a rhythm formant, corresponding to rhythmical beats of average duration 235ms, a duration known to be associated with syllables.

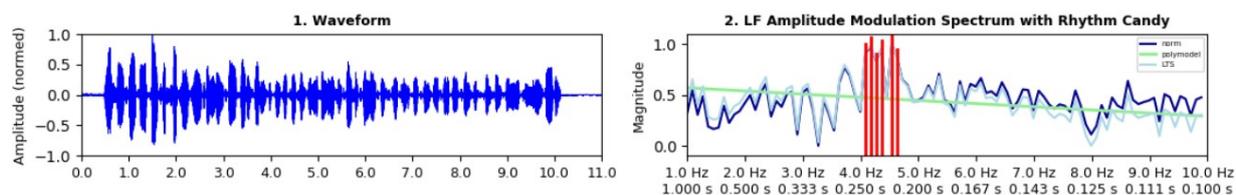

*Figure 6: Acoustic analysis of prosody parameters for a rhythmical test utterance of very rapid counting from 1 to 30. Left: time scale 0...11s for waveform and envelope. Right: frequency scale for low frequency spectrum 1...10Hz, with frequency cluster of most prominent items indicated by vertical 'rhythm bars'.*

Lexically, the numerals in the utterance represented in Figure 6 consist of monosyllables, disyllables, trisyllables and one quadrisyllable ('27'), but phonetically the very fast speech rendering results in several weak syllable deletions. Durations of all units (words, strong syllables, all syllables) tend to increase during the utterance (which can be confirmed by an analysis of an annotation of the utterance). To iron out the speech rate differences, the *nPVI* relative isochrony measure was used to measure relative syllable regularity, yielding 11 for words and 25 for strong syllables, indicating very regular timing. The *nPVI* for all syllables is 44, confirming the very regular timing compared with typical values around 60 for read-aloud English. The word count is 30, with a total duration of 9.667 s and mean duration of 322 ms, corresponding to a fast mean word rate of 3.1 per second (3.1 Hz). The mean syllable duration, based on manual annotation, is 161 ms, corresponding to a fast mean syllable rate of 6.21 per second (6.21 Hz).

The prediction based on the annotated values is that dominant frequencies in the AEMS range from about 3.1 Hz to 6.21 Hz, with a centre frequency around 4.7 Hz. Figure 6 shows the region 1 Hz to 10 Hz of the AEMS, with a cluster of 6 dominant frequencies between about 4.1 Hz and 4.6Hz. The median frequency of about 4.35 Hz differs from the predicted approximation of 4.7 Hz by only 0.35 Hz, an informal corroboration of the prediction.

The relevance of the corpus-based inductive method for language comparison and typology may be shown by the prediction that Mandarin tends, in traditional terms, to be more 'syllable-timed', while British English tends to be more 'foot-timed' or 'stress-timed', and that this prediction should be fulfilled using the inductive method. For this purpose, R-formant analyses of moderately fast fluent counting from one to thirty by a native speaker of Mandarin and a native speaker of British English are shown in Figure 7 and Figure 8, respectively, in which the 15 highest magnitude frequencies in the spectrum segment 1 Hz to 10 Hz are highlighted.

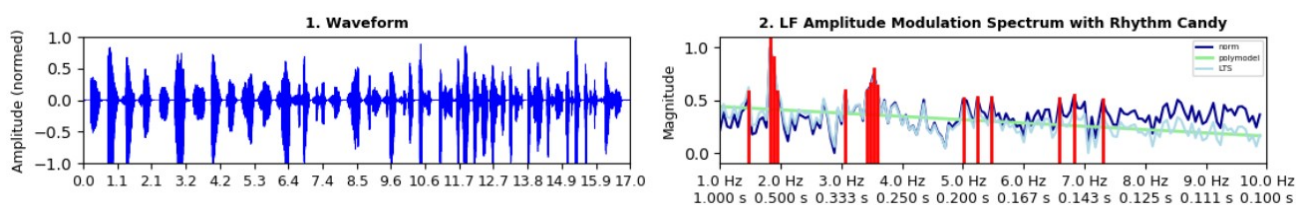

*Figure 7: Calibration of rhythm formant induction: moderately fast fluent counting 1...30 in Mandarin.*

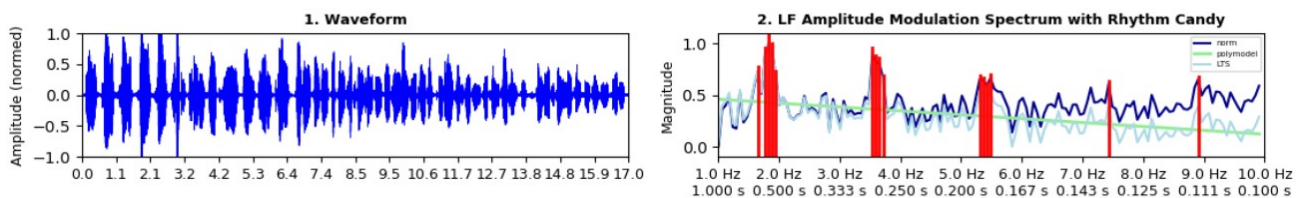

*Figure 8: Calibration of rhythm formant induction: moderately fast fluent counting 1...30 in British English.*

Figure 7 (Mandarin) shows two strong rhythm formants, at 1.9 Hz, corresponding to a word or foot rhythm, and at 3.5 Hz, corresponding to a syllable rhythm with syllable duration averaging approximately 280 ms. Other frequencies are isolated effects of non-systematic syllable shortening (above 5 Hz) or slight hesitation (the lower frequency of about 1.5 Hz).

In contrast, Figure 8 (British English) shows three strong rhythm formants, two of them with the same values as in Mandarin and a third at 5.5 Hz, corresponding to the systematically occuring weak syllables which are characteristic of British English foot or stress timing. The figures show that both Mandarin and English have foot timing, but that the essential difference is that English has a very clear distinction between systematically strong and weak syllables, while Mandarin has strong syllables and weaker patterns of non-systematic syllable weakening which do not amount to clear rhythm formants.

The analyses are implemented in Python with the libraries NumPy, MatPlotLib, and SciPy (Gibbon 2019). Display and analysis parameters are set in a configuration file. The code is available in the GitHub online portal[1] (cf. also Section 9.4, Code Appendix). An online version is also available for practical teaching purposes.[2]

## 5.2 Correlation of amplitude and frequency modulation of speech

Rhythm is a complex function of many production and perception factors, and the measurement of rhythm in the acoustic domain is subject to many decisions, some *ad hoc*, concerning parameters of signal processing. One of the basic problems to resolve in the acoustic domain is the relation between the amplitude modulation factor and the frequency modulation factor: how much does AM and how much does FM contribute to speech rhythm?

A full examination would involve, first, an examination of the perceived rhythmicity of the utterances concerned, second, measurement of the correlation of AEM spectra and FEM spectra with these judgements and, third, correlation of the AM and FM spectra with each other. Resources

---

for testing perceived rhythmicity were not available for the present study, so, as a first step the correlations of AEM and FEM spectra were examined.

An exploratory pilot experiment was carried out for this purpose. The primary data are 10 female and 10 male native speakers of standard Mandarin, reading a Mandarin translation of the IPA benchmark text, the Aesop fable *The North Wind and the Sun*. The utterances are between 40 and 60 seconds long. Initial and final silences were cropped. Secondary data for the purpose of initial informal comparison consist of a reading of the English version of *The North Wind and the Sun* by the late David Abercrombie, from the Edinburgh speech archive, and of two renderings of rhythmical counting in English: a fast speech rending of the sequence one to thirty (cf. The illustration in Figure 6) and a slow speech rendering of the sequence from one to ten (cf. Figure 9).

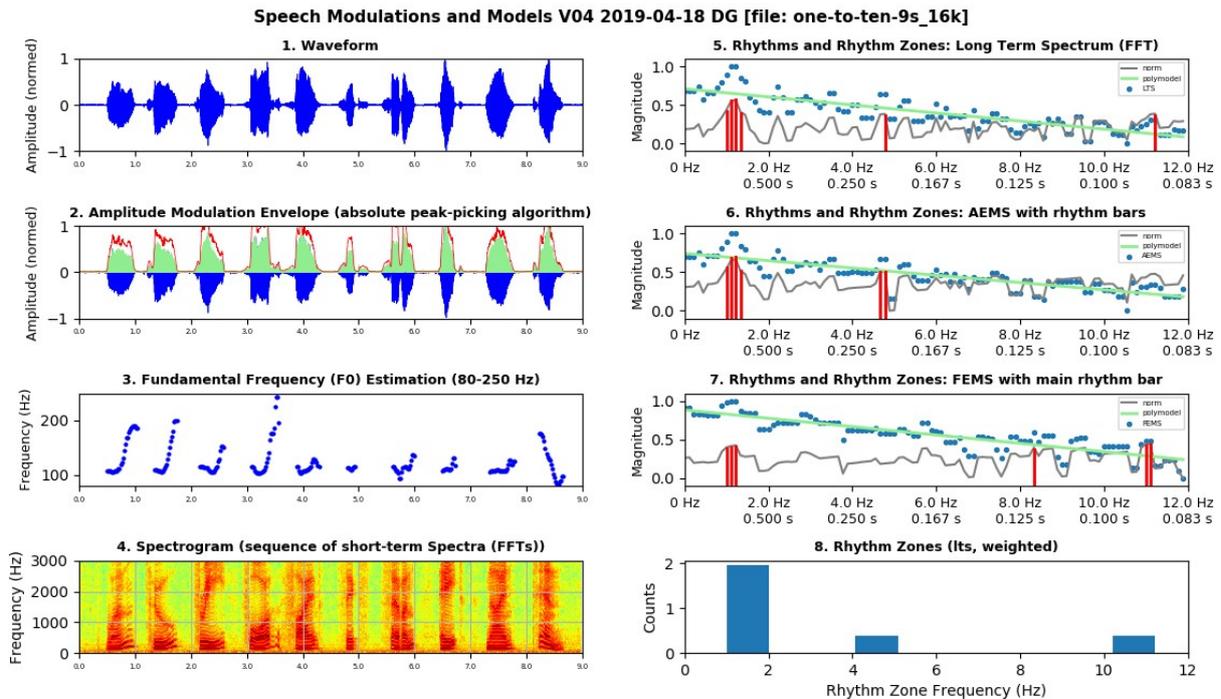

*Figure 9: Rhythm Formant Theory Analysis of an utterance with counting from one to ten. First row: waveform and rectified waveform LFS. Second row: AEM and AEMS. Third row: FEM (F0, 'pitch') and FEMS. Fourth row: spectrogram and R-formant histogram (abstraction over AEMS). (Figure as generated on screen; text and numerals are less important than shapes.*

The analysis steps for rhythmical slow counting from one to ten are visualised in Figure 9: row 1, the waveform and the raw amplitude modulation spectrum; row 2, the amplitude envelope of the waveform and the associated spectrum; row 3, the frequency modulation (F0, 'pitch') pattern and the associated spectrum. Row 4 shows the spectrogram and a generalised representation of the R-formants in binned histogram format weighted by frequency magnitudes.

The RFA method was applied, and the spectra were collected for each utterance. The F0 parameters were adjusted to typical female and male fundamental frequency settings. The lengths of the spectra were dynamically aligned, since the amplitude envelope spectrum and the frequency spectrum have different lengths. Pearson's *r* was calculated for the aligned AEMS and the FEMS spectra in order to gain a rough measure of similarity of the shapes of the spectra.

The bar chart in histogram format in Figure 9 shows a dominant R-formant between approximately 1 Hz and 2 Hz, indicating a rhythmical sequence of units of approximately 1 s duration, which can be immediately verified as the numeral words by examination of the envelope pattern of the waveform. In this slow counting style, the individual numbers have the status not only of syllables but also of words and phrases, showing list information which is extremely regular from syllable-word-phrase to syllable-word-phrase. The AEMS (and the AMS) also show a small formant at about 5 Hz, reflecting syllable internal structure as well as the syllable structure of the word *seven*, which is plainly visible in the waveform. The FEMS also has a strong R-formant at about 1 Hz, aligning closely with the R-formants of the amplitude spectra. In addition, there are small FEMS R-formants at about 8.5 Hz and 11.5 Hz, reflecting rapid intra-syllabic F0 changes at the stress positions.

One prediction is that AEMS:FEMS correlations differ in different genres: lower in read-aloud speech than in rhythmical counting, and varying with utterance length. Another prediction is that correlations differ between a tone language like Mandarin and a stress language like English on grammatical grounds, on the assumption that lexical tones, and therefore F0, vary relatively arbitrarily from syllable to syllable and therefore do not correlate strongly, whereas lexical stresses tend to be expressed with a fairly constant F0 pattern over stress groups in any given utterance and will therefore correlate relatively well.

*Table 4: Pearson's r between AM spectrum and FM spectrum*

| ID | Language | Gender | AEMS:FEMS | Median | Mean |
|----|----------|--------|-----------|--------|------|
| A | Mandarin, NWaS | F | 0.64 | | |
| B | Mandarin, NWaS | F | 0.58 | | |
| C | Mandarin, NWaS | F | 0.28 | | |
| D | Mandarin, NWaS | F | -0.04 | | |
| E | Mandarin, NWaS | F | 0.12 | 0.28 | 0.32 |
| F | Mandarin, NWaS | M | 0.81 | | |
| G | Mandarin, NWaS | M | 0.43 | | |
| H | Mandarin, NWaS | M | 0.10 | | |
| I | Mandarin, NWaS | M | -0.34 | | |
| J | Mandarin, NWaS | M | 0.07 | 0.10 | 0.21 |
| | *Overall median/mean A-J:* | | | 0.20 | 0.27 |
| DG | English 1-10 | M | 0.94 | | |
| DG | English 1-30 | M | 0.27 | | |
| DA | English NWaS | M | 0.69 | | |

The results of the experiment are shown in Table 4 and show a clear difference between Mandarin and English. The correlations for the Mandarin speakers vary very strongly, but the trend is for Mandarin amplitude modulation and frequency modulation correlation to be much lower than for English for readings of the story *The North Wind and the Sun*. This low AM:FM correlation for Mandarin was expected on functional grounds: the fundamental frequency patterns are lexically, not phrasally determined, while the opposite is true for English: the shape and frequency of pitch accents are phrasally and not lexically determined.

The Mandarin results are very different on average from the English result, though the highest outlier for a Mandarin speaker is close to the correlation for an English speaker, evidence for language internal and speaker-specific variability. Table 4 therefore also suggests that not only language but also genre and gender may be specific factors in rhythm variation, and that homogeneous descriptions claimed for entire languages are overly bold enterprises.

An clear topic for further analysis is suggested by the result that AEMS:FEMS correlations for Mandarin male speakers appear to be tendentially the lowest, followed by correlations for Mandarin female speakers. Correlations for the English speakers are much higher overall, but again, these individual results provide no more than a hint that the relative homogeneity of pitch realisations of stress patterns in an intonation language may support higher AEMS:FEMS correlations.

It is also intuitively clear that in addition to language differences there are genre differences: the highest correlation, for the short utterance with counting from one to ten, results from a higher level of homogeneity than could be expected for longer utterances. The next highest correlation is found in the longer duration of counting from one to thirty, while the lowest AEMS:FEMS correlation for English is found in the considerably longer analysis of the reading of *The North Wind and the Sun*.

### 5.3 Variation in R-formant patterns

The large variation in correlation values indicates considerable differences both in AM or in FM spectra. The weighted binned histograms of the highest magnitude spectral frequencies shown in Figure 10 (here defined on the rectified, i.e. absolute, signal rather than as AEMS formants) demonstrate that the R-formants of the Mandarin speakers pattern show some regularities, but also variation in individual speaking styles, accounting for the broad dispersion of the correlation results in Table 4.

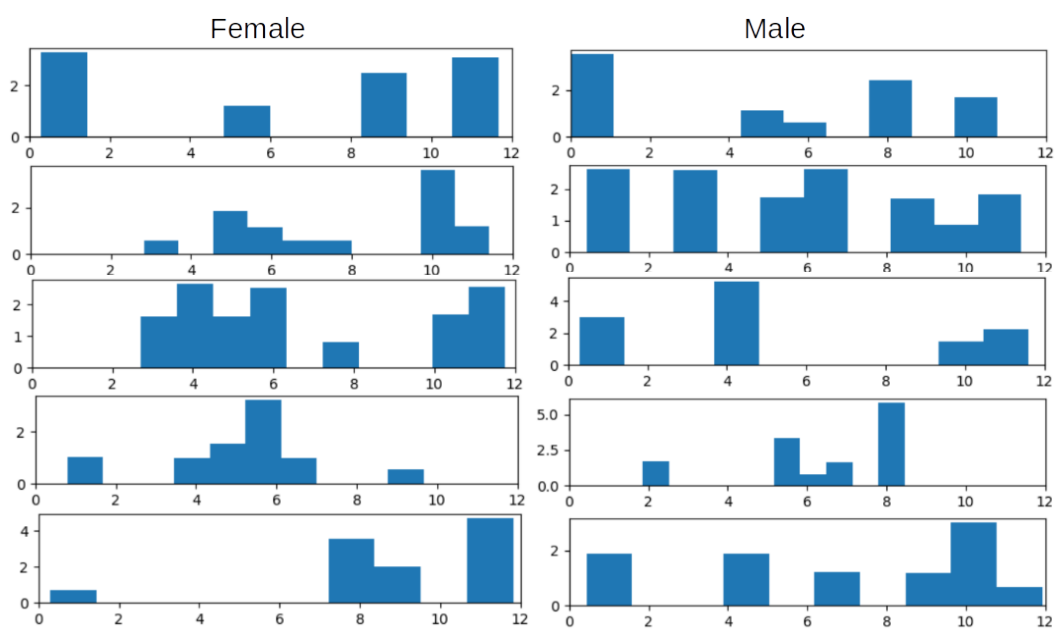

*Figure 10: Weighted binned R-formant patterns for Mandarin readers in the spectral range 0Hz to 12Hz.*

On further inspection there are some tentative generalisations which can be made. Most of the readers have conspicuous R-formants at between 4 Hz and 8 Hz, as expected, corresponding to sequences of the common syllable duration between 250 ms and 125 ms, respectively. A shift of the R-formants in this range to a lower frequency indicates a slower speaker, and a shift to a higher frequency indicates a faster speaker, in terms of syllable rate. R-formants in the range between 10 Hz and 12 Hz indicate shorter units, such as weaker and shorter syllables (for example grammatical items with the Mandarin neutral tone). Particularly interesting is the presence of R-formants below 1.5 Hz, corresponding to rhythmic patterns of phrases and larger discourse units, a topic addressed in conversation analysis (cf. Couper-Kuhlen 1993) but not in grammatical studies of prosodic phonology. Whether these units are indicated by pauses, by changes in the durations of smaller units, or in the tempo of the utterance, is a matter for further investigation.

## 6    Summary and conclusion

The present study completes the defective hypothetico-deductive cycle of mainstream linguistic studies of prosodic grammar by grounding the study of prosodic timing in a language-independent analysis of the speech signal. The main aim of the present methodologically focussed exploratory contribution is to introduce a new inductive signal-phonetic approach to the empirical grounding of prosody-grammar relations. The approach, Rhythm Formant Theory (RFT) utilises an analytic methodology which specifies rhythm formants (R-formants) quantitatively as higher magnitude frequency clusters in the long term spectra (LFS) of the amplitude envelope modulation and the frequency envelope modulation of the speech signal.

The relation between deductive and inductive approaches to rhythm analysis is discussed with reference to deductive and inductive stress modelling in generative, metrical and finite state phonology. It is further discussed in relation to the deductive measurement of timing patterns in annotation-based isochrony metrics.

The inductive RFT approach is first illustrated and calibrated by demonstrating the presence of R-formants in clearly rhythmical speech, and then the correlation between R-formants in the amplitude envelope modulation spectra (AEMS) and frequency envelope modulation spectra (FEMS) of speech is discussed in detail using read-aloud data from male and female Mandarin speakers, with individual examples from English speakers for informal comparison. The results show that an informed grammatical interpretation of trends is possible, with AEMS:FEMS correlation tendentially lower in Mandarin than in English, due to the more heterogeneous FM patterns of a tone language and the more homogeneous FM patterns of a stress language. Tendential differences between male and female Mandarin speakers and genre differences in English between rhythmical counting and reading aloud were also observed.

The Mandarin data set is very small (10 speakers) and variation is considerable, and for English only individual examples are given. Therefore, clearly, statistical significance is not on the cards. The study is deliberately exploratory, not a full confirmatory investigation, and statistical significance in this stage of development is less important than linguistically informed interpretation of trends which point to possible fruitful topics for future research. Applications of the RFT methodology are anticipated in language testing, clinical speech diagnosis, language typology, naturalness evaluation in speech synthesis, and in speaker recognition.

# 7 Acknowledgments

First, thanks to the reviewers: reviewer 1 provided detailed feedback, while reviewer 2 expressed a general scepticism, a challenge which has been addressed. Special thanks go to Jue Yu, for extensive discussion and for kindly providing Mandarin corpus data, to Sheida Marzban for Farsi data, to Peng Li for Mandarin calibration data, and to Huangmei Liu for suggesting the term 'formant' for the high magnitude frequency clusters. For comments and suggestions at various points in the development of Rhythm Formant Analysis I am particularly indebted to Laura Dilley, Alexandra Gibbon, Xuewei Lin, Rosemarie Tracy, Petra Wagner and Ratree Wayland.

Special thanks also go to Qiuyu Ma, for three opportunities to contribute to the Summer Schools on "Contemporary Phonetics and Phonology" at Tongji and Fudan Universities, Shanghai, and to the students there for insightful questions and suggestions. Warmest thanks, too, to Qi Gong, for the opportunity to interact and publish with students and colleagues at the School of Foreign Studies, Jinan University, Guangzhou as Visiting Professor for for the period 2016 to 2019, during which the present approach was developed. Much of the text of the present study is taken, unabashedly and partly verbatim, but without infringement of intellectual property rights, from my own previous work and my lectures in Shanghai and Guangzhou, in acknowledgment of difficulties which some readers may have in accessing this previous work.

# 9   Code Appendix

## 9.1   First steps in computational phonology

This appendix is addressed to readers who are potentially interested in computational phonology and would like to have a starting point for experimenting with computational syntax-prosody mappings in phonology and with the RFT methodology. The first maps a list containing a string of parentheses and words, representing an expression in a parenthesis language to a string with the words index with 'stress numbers'. The second is very simple and, ignoring the words, maps the sequence of numbers to a string in the parenthesis language. The third is complex, for the intermediate level Python user, and enables visualisation of R-formants in WAV files.

In order to reduce space in the present context, the code requires well-formed input in order to produce well-formed output, none of the usual software precautions against semantic error are included, and no documentation comments are included. These steps are recommended as an exercise for the interested reader.

## 9.2   Nuclear stress rule: generator from parenthesis language to integer sequence

```python
#!/usr/bin/python3
# nsrbrackets.py
# D. Gibbon, 2019-09-13

# Initialise input bracketing
bracketing = "( ( tiny Moll ) ( met ( tall Jill ) ) )"
bracketing = bracketing.split(" ")

# Initialise vocabulary, counter and output variables
leftbracket = "("
rightbracket = ")"
brackets = [ leftbracket, rightbracket ]
counter = 1
output = []
lastitem = "init"

# Iterate through input bracketing
for item in bracketing:
    if item == leftbracket:
        if lastitem == rightbracket:
            output = output[:-1] + [str(counter) + output[-1]]
        counter += 1
    if not item in brackets:
        if  lastitem == leftbracket:
            output += [ str(counter) + item ]
        else:
            output += [ item ]
    if item == rightbracket:
        counter -= 1
    lastitem = item
```

```
# Termination
output = output[:-1] + [str(counter) + output[-1]]
print("Output:", output)
```

## 9.3 Inverse nuclear stress rule: simple parser from integers to parenthesis language

```
#!/usr/bin/python3
# nsrbracketsinverse.py
# D. Gibbon, 2019-09-13

leftbracket = "("
rightbracket = ")"
brackets = [ leftbracket, rightbracket ]
numberstring = "3 4 2 3 4 1"
numbers = list(map(int, numberstring.split(" ")))
counter = 1
output = []
lastitem = "init"
output = ""
lastitem = 1

for item in numbers:
        if lastitem < item:
                output += "("*(item-lastitem) + " " + str(item) + " "
        if lastitem > item:
                output += " " + str(item) + " " + ")"*(lastitem-item)
        lastitem = item

print("Inverse nuclear stress rule:")
print("Input:", numberstring)
print("Output:", output)
```

## 9.4 R-formant visualisation

```
#!/usr/bin/python3
# aems.py
# D. Gibbon, 2019-09-15
# Visualise waveform, amplitude envelope modulation AEM
# and low-frequency spectrum AEMS (requires WAVE filename)

import sys, re
import numpy as np
import matplotlib.pyplot as plt
import scipy.io.wavfile as wave
from scipy.signal import medfilt

wavfilename = sys.argv[1]

fontsize = 9
graphwidth = 12
graphheight = 3
wavecolor = "blue"
abswavecolor = "lightblue"
envcolor = "r"
graphformat = "png"
showgraph = True
envwin = 20
envmedianfilt = 501
envheight = 1.1
```

```
rhythmzones = "aems"
rhythmcount = 6
spectrumpower = 2
spectrumpoly = 1
aemsmin = 1
aemsmax = 10
aemsmedfilt = 3

fs, signal = wave.read(wavfilename)
if len(signal.shape) > 1: signal = (signal[:,0]/2.0 + signal[:,1]/2.0)
signal = signal/float(max(abs(signal)))
signallen = len(signal)
signalsecs = float(signallen)/fs
signalstart = 0
signalend = signalsecs

def makeenvelope(signal, envwin, envmedianfilt):
        signalabs = abs(signal)
        peaksrange = range(len(signalabs)-envwin)
        envelope = [ max(signalabs[i:i+envwin]) for i in peaksrange ]
        padleft = [envelope[0]] * int(round(envwin/2.0))
        padright = [envelope[-1]] * int(round(envwin/2.0))
        envelope = padleft + envelope + padright
        envelope = np.asarray(envelope)
        envelope = medfilt(envelope,envmedianfilt)
        envelope = envelope / float(max(envelope))
        return np.asarray(envelope)

def fft(signal,fs):       # signal is a NumPy array; fs is sampling rate integer.
        period = 1.0/fs
        mags = np.abs(np.fft.rfft(signal))**2
        freqs = np.abs(np.fft.rfftfreq(signal.size,period))
        mags = np.asarray([0.000001 if m==0 else m for m in mags])
        mags = np.log10(mags)    # CHECK IF THIS IS NECESSARY
        mags[0] = mags[1]
        mags = mags / np.max(mags)
        return freqs, mags

def polyregline(x,y,d):
        x = range(len(y))
        fit, res, _, _, _ = np.polyfit(x, y, d, full=True)
        yfit = np.polyval(fit,x)
        return yfit

envelope = makeenvelope(signal, envwin, envmedianfilt)
aemsf0,aemsmags = fft(abs(envelope),fs)
aemsmags[0] = aemsmags[1]
aemsmags = medfilt(aemsmags,aemsmedfilt)

# Initialise graph
_,(    pltenv, pltaems
        ) = plt.subplots(nrows=1, ncols=2, figsize=(graphwidth, graphheight))

# Amplitude envelope
title = "AEM (absolute peak-picking algorithm)"
pltenv.set_title(title)
width = 2.0; ymin = -1.0; ymax = 1.0
x = np.linspace(signalstart,signalend,signallen)
pltenv.plot(x,signal, color=wavecolor, linewidth=width)
```

```
pltenv.plot(x,abs(signal), color=abswavecolor, linewidth=width)
pltenv.plot(x,envheight * abs(envelope), color=envcolor, linewidth=width)
xx = np.linspace(signalstart,signalend,signalsecs + 1)
pltenv.set_xticks(xx)
xtix = pltenv.get_xticks()
xtixlabels = [ "%.1f"%(i) if i>0 else str(signalstart) for i in xtix ]
pltenv.set_xticklabels(xtixlabels,fontsize=fontsize)
pltenv.set_ylim(ymin, ymax)
pltenv.set_xlim(signalstart, signalend)

# Long term amplitude envelope spectrum

title = "Rhythms and Rhythm Zones: AEMS with rhythm bars"
pltaems.set_title(title)
ymin = 0
ymax = 1.0
width = 2.0
xminsamples = int(aemsmin * fs/2.0)
xmaxsamples = int(aemsmax * fs/2.0)
numin = int(round(aemsmin * len(aemsf0) / aemsf0[-1]))
numax = int(aemsmax * len(aemsf0) / aemsf0[-1])
aemsf0 = aemsf0[numin:numax]
data = aemsmags[numin:numax]
data = data**spectrumpower
data = np.log10(data)
data = (data - np.min(data)) / (np.max(data)-np.min(data))
x = np.arange(len(data))
polymodel = polyregline(x,data,spectrumpoly)
polyresid = data - polymodel
polyresidnorm = polyresid+abs(np.min(polyresid))
pltaems.plot(aemsf0,polyresidnorm, color='b', label='norm')
if rhythmcount > 0:
        speclist = polyresidnorm.tolist()
        speclist = speclist
        speclistsort = sorted(speclist)
        speclistsortrev = reversed(speclistsort[-rhythmcount:])
        for i, item in enumerate(speclistsortrev):
                b = speclist.index(item)        # actual vector position
                f = float(b)/signalsecs # vector position in Hz
                print(i,item,f)
                f = f + aemsmin                 # if signal does not start at zero
                pltaems.axvline(f,ymin=0, ymax=item, linewidth=2, color='r')
pltaems.set_xlim(aemsmin, aemsmax)
pltaems.set_ylim(ymin-0.1, ymax+0.1)
xtix = pltaems.get_xticks()
xtixlabels = [ "%.1f Hz\n%.3f s"%(i,1.0/i) if i>0 else "0 Hz" for i in xtix ]
pltaems.set_xticklabels(xtixlabels,fontsize=fontsize)
pltaems.set_ylabel("Magnitude", fontsize=fontsize)

plt.tight_layout(pad=3, w_pad=1, h_pad=1)
plt.savefig("AEMS.png")
if showgraph:
        plt.show()
```